\documentclass[conference]{IEEEtran}
\IEEEoverridecommandlockouts
\usepackage{cite}
\usepackage{amsmath,amssymb,amsfonts}
\usepackage{graphicx}
\usepackage{textcomp}
\usepackage{xcolor}
\usepackage{multirow}
\usepackage{amsmath}
\usepackage{amsthm}
\usepackage{algorithm}
\usepackage{algpseudocode}
\usepackage{url}
\PassOptionsToPackage{hyphens}{url}
\usepackage{hyperref}
\usepackage{makecell}
\usepackage{multirow}
\usepackage{graphicx}
\usepackage{wrapfig}
\graphicspath{ {images/} }
\usepackage{color}

\usepackage[caption=false,font=footnotesize]{subfig}

\def\BibTeX{{\rm B\kern-.05em{\sc i\kern-.025em b}\kern-.08em
    T\kern-.1667em\lower.7ex\hbox{E}\kern-.125emX}}
\begin{document}

\title{Rethinking Learning Rate Tuning in the Era of Large Language Models}

\author{\IEEEauthorblockN{Hongpeng Jin$^*$, Wenqi Wei$^\dagger$, Xuyu Wang$^*$, Wenbin Zhang$^*$, Yanzhao Wu$^*$}
\IEEEauthorblockA{\textit{$^*$ Florida International University, Miami, FL 33199} \\
\textit{$^\dagger$ Fordham University, New York City, NY 10023}\\
}
}

\maketitle

\begin{abstract}
Large Language Models (LLMs) represent the recent success of deep learning in achieving remarkable human-like predictive performance. It has become a mainstream strategy to leverage fine-tuning to adapt LLMs for various real-world applications due to the prohibitive expenses associated with LLM training. The learning rate is one of the most important hyperparameters in LLM fine-tuning with direct impacts on both fine-tuning efficiency and fine-tuned LLM quality. Existing learning rate policies are primarily designed for training traditional deep neural networks (DNNs), which may not work well for LLM fine-tuning. We reassess the research challenges and opportunities of learning rate tuning in the coming era of Large Language Models. This paper makes three original contributions. First, we revisit existing learning rate policies to analyze the critical challenges of learning rate tuning in the era of LLMs. Second, we present LRBench++ to benchmark learning rate policies and facilitate learning rate tuning for both traditional DNNs and LLMs. Third, our experimental analysis with LRBench++ demonstrates the key differences between LLM fine-tuning and traditional DNN training and validates our analysis.
\end{abstract}


\section{Introduction}
Deep learning and its applications have achieved remarkable success and recognition in numerous areas, including object detection~\cite{object-detection}, self-driving cars~\cite{self-driving-car}, drug discovery~\cite{askr2023deep}, laboratory automation~\cite{self-driving-lab}, speech recognition~\cite{radford2023robust}, and chatbot~\cite{chatgpt, bard, claude, llama}, ranging from Edge AI~\cite{parallel-detection-cogmi2021,fast-resource-efficient-object-tracking,edge-ai-survey} to the recent success of Large Language Models (LLMs)~\cite{chatgpt, bard, claude, llama}.
LLMs can approach near-human levels of intelligence in many Natural Language Processing (NLP) tasks, such as question answering, machine translation, reading comprehension, and summarization~\cite{arc, bleu, triviaqa, coqa, superglue,gpt3}, which has inspired numerous real-world applications, such as ChatGPT~\cite{chatgpt}, Claude~\cite{claude}, Bard~\cite{bard}, and LLaMA~\cite{llama}.
The recent rapid development of LLMs represents a new wave of AI technology innovations and extensive impacts, bringing the world into a new era of Large Language Models.

LLMs leverage the unprecedented model complexity (at the 100s of billions level of \#model parameters) to achieve close to human-level intelligence at very high training costs (at least \$10 million~\cite{new-approach-train-LLM-half-time}), which introduces critical challenges in training LLMs. Unlike traditional deep neural network (DNN) training which is still generally affordable, requiring only several GPUs, training LLMs from scratch has been only possible with the thousands of GPUs and huge investments from big tech companies and/or large research labs. The emergence of LLMs is changing the way deep learning models are distributed and adopted in real-world applications. 
We envision that fine-tuning LLMs will dominate the technology diffusion of newly developed LLMs in the future. Pre-trained LLMs will be released by big tech companies and/or large research labs as foundation models. Application developers and academic researchers will fine-tune pre-trained LLMs to develop real-world applications or study domain-specific research problems. 
However, it lacks systematic studies on how to achieve high efficiency and high accuracy in LLM fine-tuning. Most existing LLM fine-tuning methods still follow similar assumptions/principles to traditional deep learning training/fine-tuning~\cite{gpt1,instructgpt,alpaca,vicuna,koala}. However, the high model complexity and fine-tuning costs of LLMs make it a pressing challenge that demands the reassessment of these fundamental assumptions/techniques for LLM fine-tuning.
For example, LLM fine-tuning and traditional DNN training both involve multiple hyperparameters, such as the learning rate policy and batch size~\cite{vgg,resnet,gpt1,llama}. The hyperparameter tuning strategies for LLM fine-tuning are primarily based on the experience or practice obtained from traditional DNN training, which may not work well for LLM fine-tuning.

The learning rate (LR) is one of the most critical hyperparameters and directly impacts both the DNN training effectiveness and the trained model accuracy, which plays a similar role in LLM fine-tuning.
LLM fine-tuning can be formulated as an iterative optimization problem to minimize a pre-defined loss function $L$, where an optimizer, such as Adam~\cite{adam}, will update the LLM model parameters $\Theta$ using the learning rate $\eta(t)$ and gradients $\nabla L$ for the iteration $t$, following $\Theta_{t+1} = \Theta_t - \eta(t) \frac{\hat{M}_t}{\sqrt{\hat{V}_t} + \epsilon}$. $\hat{M}_t$ and $\hat{V}_t$ are exponential moving averages of gradient $\nabla L$ and squared gradient $(\nabla L)^{2}$.
The learning rate directly controls the magnitude of gradients to be updated on the pre-trained LLM, which allows the optimizer to adjust the learning speed for each iteration.
However, it can be a daunting task to find a good learning rate. Too small or too large learning rates will impair LLM fine-tuning, leading to either failure in model convergence or suboptimal performance. Moreover, the conventional trial-and-error approach leverages manual tuning to try different learning rates each time, which is tedious, time-consuming, and highly expensive for LLMs.
Similar challenges are also being faced by traditional deep learning training~\cite{optuna,dawnbench,ray-tune,GTDLBenchICDCS,GTDLBenchBigData,GTDLBenchTSC}.

Therefore, it has been an open challenge in both traditional DNN training and LLM fine-tuning to identify good learning rate policies. In the era of Large Language Models, there still lacks a systematic study of learning rate tuning to determine whether the high complexity and unique characteristics of LLMs bring in new research challenges and/or opportunities.
We examine and answer these questions in this paper and make three original contributions.
{\it First,} we revisit and summarize existing learning rate policies for improving traditional deep learning training.
{\it Second,} we analyze the fundamental assumption changes for LLMs and identify the key research challenges and opportunities in the upcoming era of Large Language Models.
{\it Third,} we perform experimental evaluations using LRBench++ for both traditional DNN training and LLM fine-tuning, which confirms our analysis and reveals future research directions.

\section{Related Work}
We below summarize related studies from three main perspectives, (1) learning rate tuning, (2) hyperparameter tuning, and (3) Large Language Model (LLM) fine-tuning.

\noindent \textbf{Learning Rate Tuning.}
The learning rate has been widely recognized as one of the most important hyperparameters with high impacts on traditional DNN training~\cite{clr,superconvergence,largeminibatch,understanding-lr-blog,LRBenchBigData,LRBenchTIST,autolr,autolrs}. Existing studies can be summarized into three broad categories of learning rate policies.
(1) \textit{Formula-based LR} is one of the most popular methods to specify learning rate values by using a pre-defined function ($\eta(t)$) of the training iterations/epochs ($t$)~\cite{clr,superconvergence,sgdr,LRBenchBigData}. These formula-based LRs are summarized by~\cite{LRBenchBigData,LRBenchTIST} into fixed LR, decaying LR, cyclic LR, and composite LR.
(2) \textit{State-based LR} is represented by Reduce-LR-On-Plateau~\cite{reducelronplateau} and Change-LR-On-Plateau~\cite{LRBenchTIST}, where the learning rate values are computed on-the-fly based on the deep learning training state, such as the training/validation loss. 
(3) \textit{Exploration-based LR} models learning rate tuning with Reinforcement Learning~\cite{rl-lr-tuning}, Bayesian optimizations~\cite{autolrs}, or evolutionary approaches~\cite{autolr,leaderpopulation}, to explore the promising search space of learning rates and identify good values in deep learning training.

\noindent \textbf{Hyperparameter Tuning}. This thread of efforts aims to provide a general-purpose hyperparameter tuning framework, such as Ray Tune~\cite{ray-tune}, Hyperopt~\cite{hyperopt}, SMAC~\cite{smac} and Optuna~\cite{optuna}, by using various tuning strategies, such meta modeling~\cite{metamodeling-ht}, bayesian optimization~\cite{bayesianoptimization-ht}, bandit-based approach~\cite{banditbased-ht}, and population-based method~\cite{populationbased-ht}. Even though learning rate can be considered as a sub-problem of hyperparameter tuning, existing general-purpose hyperparameter tuning frameworks lack dedicated support for learning rate tuning, such as tuning various decaying/cyclic LRs or even composite LRs~\cite{LRBenchTIST}.

\noindent \textbf{Large Language Model (LLM) Fine-tuning.} Fine-tuning LLMs is an efficient and practical method to adapt a pre-trained LLM to a new learning task, which requires much lower training costs and less data compared to training LLMs from scratch~\cite{alpaca,koala,vicuna}. 
Alpaca~\cite{alpaca}, Koala~\cite{koala}, and Vicuna~\cite{vicuna} are representative fine-tuned LLMs based on the public release of LLaMA~\cite{llama}, which exhibit improved conversational abilities. \cite{vicuna} also introduces two LLM benchmarks, Chatbot Arena, which is a rating system to rank LLMs based on crowdsourced user votes, and MT-bench, which leverages another LLM as a judge to evaluate the performance of LLMs.

To the best of our knowledge, this study is the first to explore learning rate tuning for LLMs.

\section{Problem Statement}

\begin{figure}[h!]
    \centering
    \includegraphics[width=0.5\textwidth]{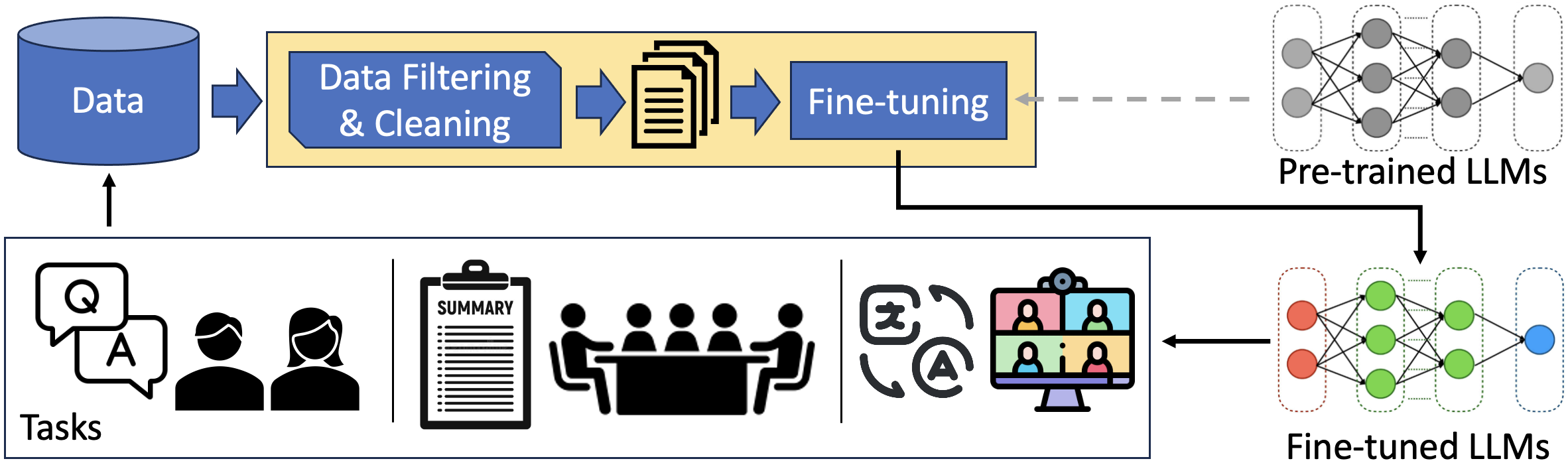}
\caption{The LLM Fine-tuning Workflow}
\label{fig:fine-tuning-pipelines}
\end{figure}

The LLM fine-tuning is a process that optimizes a pre-trained LLM to improve the predictive performance on a new dataset and/or a new learning task. 
Figure~\ref{fig:fine-tuning-pipelines} shows the workflow of LLM fine-tuning. 
{\it First,} we need to choose a pre-trained LLM to initiate fine-tuning, like LLaMA~\cite{llama} or Falcon\cite{falcon}, and prepare the fine-tuning data, which contains the domain knowledge for a new application or a new learning task, such as customer service materials for LLM-powered customer services and meeting summaries for meeting summary generation using LLMs~\cite{instructgpt,finetuningblog2,finetuningblog3,finetuningblog4}. 
{\it Second,} the pre-trained LLM will be fine-tuned to achieve enhanced performance and deployed to support this new application or new learning task.
{\it Third,} once deployed, these fine-tuned LLMs can generate new data, which can be filtered and cleaned to continuously improve LLM performance.

The LLM fine-tuning process involves multiple steps to leverage a loss function ($L$), an optimizer ($\mathcal{O}$), and a learning rate policy ($\eta(t)$) to iteratively optimize pre-trained LLMs.
{\it First,} the LLM will perform inference on an input batch of fine-tuning data to generate predictions.
{\it Second,} the loss function will be leveraged to compute the LLM gradients based on the difference between the prediction and ground truth.
{\it Third,} the optimizer and learning rate policy will jointly update LLMs by controlling the gradients applied to the model parameters. For example, Adam~\cite{adam} is a popular optimizer for LLM fine-tuning~\cite{alpaca,koala,vicuna}, which follows Formula~(\ref{formula:adam}) to perform model parameter updates.
\begin{equation}
\begin{aligned}
M_t &= \beta_1 M_{t-1} + (1-\beta_1) \nabla L, \quad \hat{M_t} = \frac{M_t}{1-\beta_1^t} \\
V_t &= \beta_2 V_{t-1} + (1-\beta_2) (\nabla L)^2, \quad \hat{V_t} = \frac{V_t}{1-\beta_2^t} \\ 
\Theta_{t+1} &= \Theta_{t} - \eta(t)\frac{\hat{M_t}}{\sqrt{\hat{V_i}}+\epsilon}
\end{aligned}
\label{formula:adam}
\end{equation}
where $\nabla L$ is the gradients calculated for the current iteration, $\hat{M}_t$ and $\hat{V}_t$ are exponential moving averages of gradient and squared gradient, $\beta_1$ and $\beta_2$ are the two coefficients to control the impacts of the previously accumulated gradients and square of gradients with $\beta_1=0.9$, $\beta_2=0.999$ and $\epsilon=10^{-8}$ by default, and $\beta_1^t$ and $\beta_2^t$ represent $\beta_1$ and $\beta_2$ to the power of $t$ respectively.
The above steps will repeat throughout the entire fine-tuning data (i.e., one epoch) and continue with the next epoch until reaching the pre-defined number of epochs.

Given a pre-trained LLM $F_\Theta$ with trainable model parameters $\Theta$ and fine-tuning data $\mathcal{X}^{ft}$, the optimizer $\mathcal{O}_{\eta(t)}$ with a learning rate policy $\eta(t)$ minimizes the loss $L(x;F_\Theta)$ over the fine-tuning data to produce a fine-tuned LLM with $\Theta = \mathcal{O}_{\eta(t)}(\mathcal{X}^{ft})$.
The learning rate tuning aims to find a good learning rate policy $\hat{\eta}(t)$ to effectively minimize the generalization error $\mathbb{E}[T; F_{\mathcal{O}_\eta(t)(\mathcal{X}^{ft})})]$ of fine-tuned LLMs on a specific learning task ($T$), such as question answering. This optimization problem can be formulated as follows:
\begin{equation}
\hat{\eta}(t) =  argmin_{\eta(t) \in \mathcal{P}}\mathbb{E}[T; F_{\mathcal{O}_\eta(t)(\mathcal{X}^{ft})})]
\label{formula:lr-optimization}
\end{equation}
where $\mathcal{P}$ is the set of all possible LR policies.

\begin{figure}[h!]
    \centering
    \includegraphics[width=0.5\textwidth]{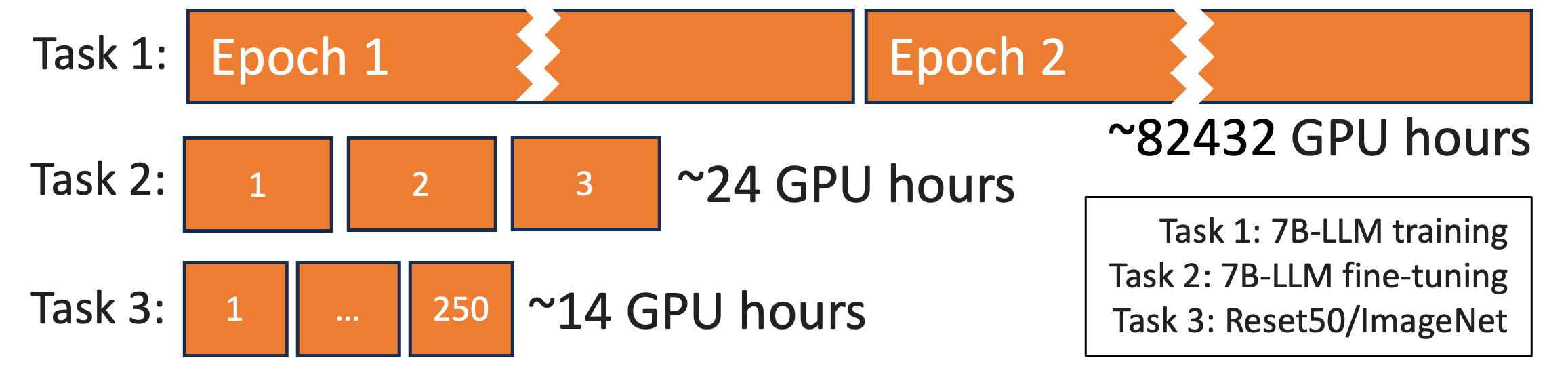}
\caption{DNN Training v.s. LLM Training/Fine-tuning}
\label{fig:training-comparasion}
\end{figure}

Even though LLM fine-tuning follows a similar paradigm to traditional deep learning training~\cite{resnet,GTDLBenchICDCS,gpt1,llama}, training/fine-tuning a large language model significantly differs from training a traditional relatively small-scale deep neural network in several aspects, including
(1) very high model complexity, where LLMs typically have over billions of model parameters while traditional DNNs only have millions of model parameters~\cite{model-zoo-pytorch,model-zoo-caffe},
(2) expensive training/fine-tuning costs, such high model complexity of LLMs makes it costly to train or even fine-tune LLMs. For example, Figure~\ref{fig:training-comparasion} shows the comparison between traditional DNN training and LLM training/fine-tuning costs in terms of \#GPU hours (NVIDIA A100-80GB)~\cite{llama,reset50-training-time}. 
Training traditional DNNs on ImageNet~\cite{ILSVRC,reset50-training-time} used to be considered a costly task before with over 10 GPU hours while the cost of LLM training/fine-tuning is much higher. In particular, training LLMs from scratch is prohibitively expensive with over 80k GPU hours (over 9 years), making it an unattainable objective without substantial efforts and investments,
(3) different model initialization, where pre-trained models are used for LLM fine-tuning while traditional DNN training often starts from random initialization~\cite{initialization-momentum,initialization-architecture},
(4) fewer training epochs, where LLM training/fine-tuning only requires 2$\sim$3 long epochs (see Table~\ref{table:fine-tuning-method-comparasion}) while traditional DNN training often requires over 100 epochs, e.g., 250 epochs for training ResNet50 on ImageNet~\cite{reset50-training-time},
and (5) different evaluation strategies, where traditional DNNs can often be quickly evaluated on testing data while LLM evaluation may involve time-consuming real-world tasks, such as taking real exams and/or being judged by another LLM~\cite{mmlu,vicuna}, making it very hard to monitor the LLM quality during training/fine-tuning.
We found that simply following the traditional DNN training/fine-tuning strategies may not work well for LLM training/fine-tuning, leading to suboptimal solutions and/or unnecessary waste of computational resources.
For example, the popular cyclic LRs in DNN training recommend a step size of 4$\sim$6 epochs while LLM training/fine-tuning only has 2$\sim$3 epochs~\cite{alpaca,vicuna}.
We argue that there is a pressing need to rethink the LLM training/fine-tuning paradigm given these key differences between traditional DNN training and LLM training/fine-tuning.

\begin{figure}[h!]
\centering
\subfloat[73th iteration]{
  \centering
  \includegraphics[trim=60 25 20 25, clip,width=0.17\textwidth]{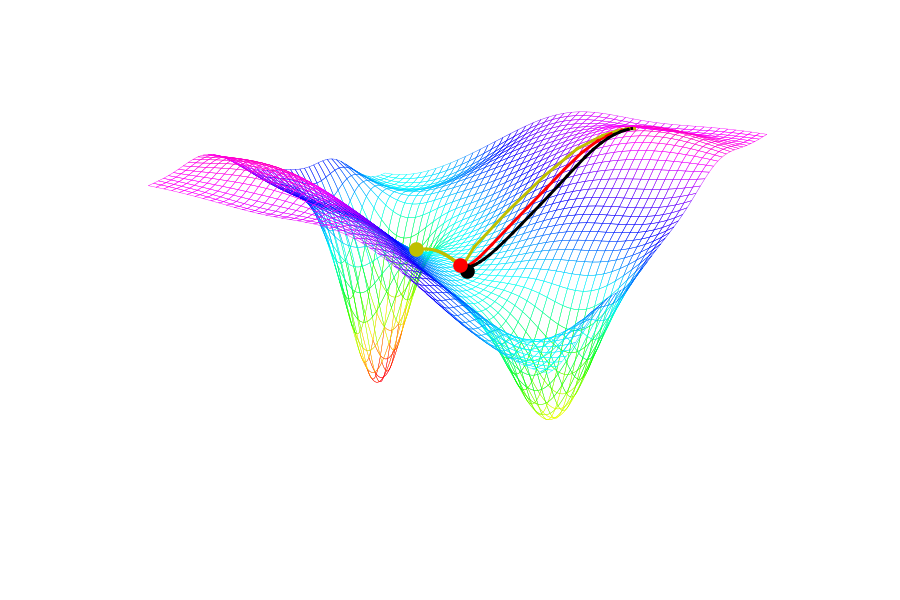}
  \label{fig:fix-nstep-triexp-73}
} 
\hspace*{-4mm}
\subfloat[110th iteration]{
  \centering
  \includegraphics[trim=60 25 20 25, clip,width=0.17\textwidth]{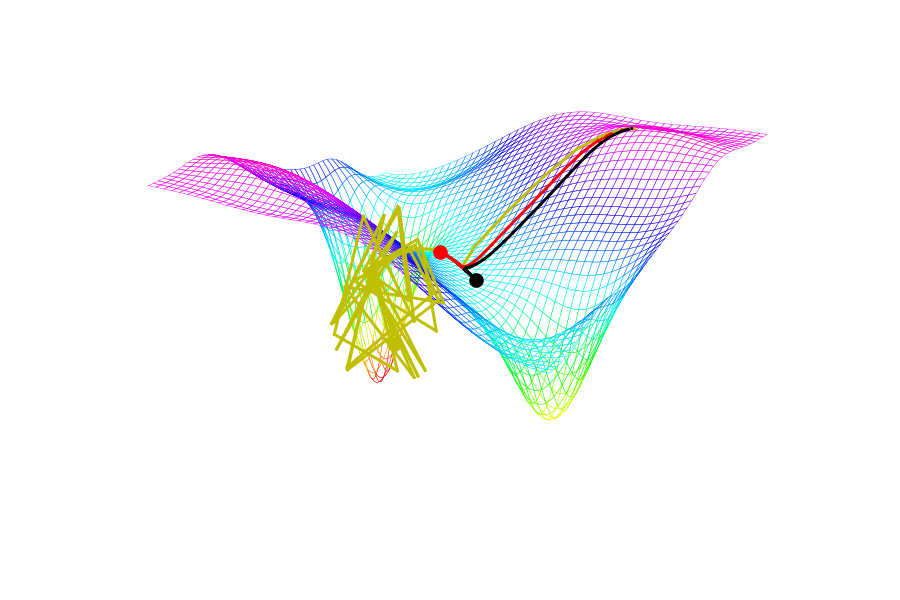}
  \label{fig:fix-nstep-triexp-110}
} 
\hspace*{-4mm}
\subfloat[140th iteration]{
  \centering
  \includegraphics[trim=60 25 20 25, clip,width=0.17\textwidth]{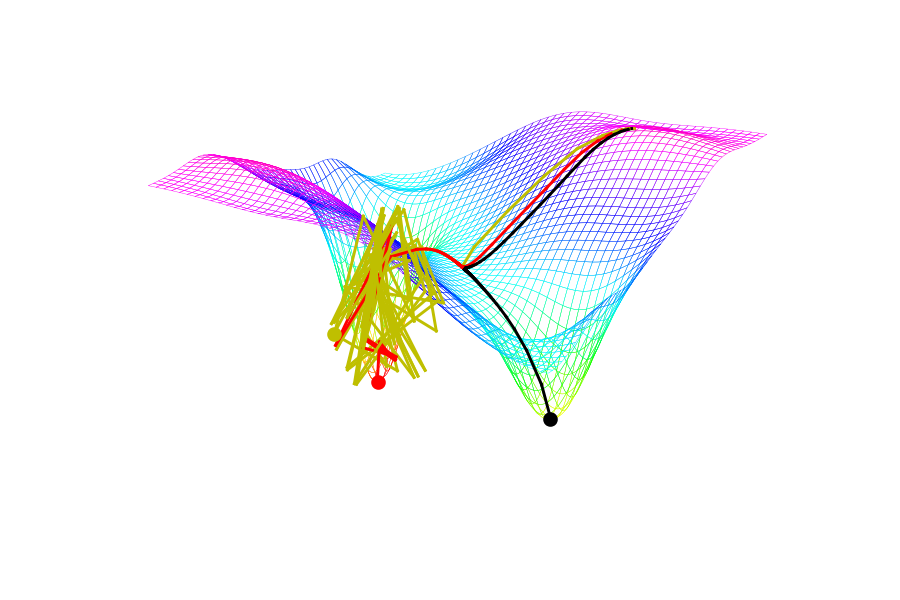}
  \label{fig:fix-nstep-triexp-140}
} 
\caption{Visualization of the Training Process with Different LRs, black: FIX($k=0.025$), red: NSTEP($k=0.05, \gamma=0.5, l=[120, 130]$), and yellow: TRIEXP($k_0=0.05, k_1=0.25, \gamma=0.9, l=25$), different LRs lead to different optimization paths}
\label{fig:visulization-fix-nstep-triexp}
\end{figure}

Learning rate tuning itself can be challenging as well. Figure~\ref{fig:visulization-fix-nstep-triexp} illustrates the optimization paths of using different LRs, FIX (black), NSTEP (red), and TRIEXP (yellow). Here, we follow the name conventions in~\cite{LRBenchBigData,LRBenchTIST} for these three different formula-based LRs.
This optimization process starts from the same initial location and lasts for 140 iterations for all three LRs. The cost function is represented using the grid, where reaching the global optimum (red color) is the objective for the optimization. In the beginning, TRIEXP is the fastest in making progress toward the global optimum in Figure~\ref{fig:fix-nstep-triexp-73}, which shows that starting with increasing the LR values ($k_0\rightarrow k_1$) may accelerate the optimization process. In the middle, FIX is approaching the wrong direction while TRIEXP still has relatively high LR values with high ``kinetic energy'' and bounces around the global optimum in Figure~\ref{fig:fix-nstep-triexp-110}. In the end, only NSTEP can reach the global optimum in Figure~\ref{fig:fix-nstep-triexp-140}. Overall, different LRs lead to different optimization paths, which indicates that the cumulative effects of LR values have a high impact on training/fine-tuning DNNs/LLMs. Hence, it is critical to identify good LRs for improving the training/fine-tuning effectiveness.
Three challenges should be addressed in finding good LRs:
(1) how to evaluate and compare the performance of different LRs and their impacts on training/fine-tuning,
(2) how to select and compose good LR policies to achieve different training/fine-tuning objectives, such as high accuracy or low cost,
and (3) how to tune LR parameters to adapt to different training/fine-tuning stages given many LR parameters and the large search space.

\section{LRBench++ Overview}

We present our solution framework, LRBench++, which is an \textbf{LR} \textbf{Bench}marking system for evaluating, selecting, and tuning LR policies to optimize DNN/LLM training/fine-tuning. LRBench++ is built on top of LRBench~\cite{LRBenchBigData}, which implements all four categories of formula-based LRs, including fixed, decaying, cyclic, and composite LRs, state-based LRs, including Reduce-LR-On-Plateau and Change-LR-On-Plateau, and evaluation metrics that measure the utility, efficiency, cost, and robustness of LR policies. 
LRBench++ augments LRBench through three major technical improvements.
{\it First,} we provide support for the training and fine-tuning of Large Language Models as well as LLM evaluation techniques to address the pressing challenges of LLM training and fine-tuning.
{\it Second,} we implement popular hyperparameter tuning algorithms, including grid search and random search, into LRBench++, which will further improve the LR tuning efficiency.
{\it Third,} we incorporate various LR tuning methods for accomplishing different DNN/LLM training/fine-tuning objectives into LRBench++, which allows it to cater to diverse real-world applications with different training/fine-tuning requirements, such as attaining high accuracy or minimizing costs.
Our ongoing work includes incorporating diverse learning tasks, such as object detection and sentiment analysis, and exploration-based LRs into LRBench++, which will further enhance its benchmarking and tuning capability.

\begin{figure}[h!]
    \centering
    \includegraphics[width=0.5\textwidth]{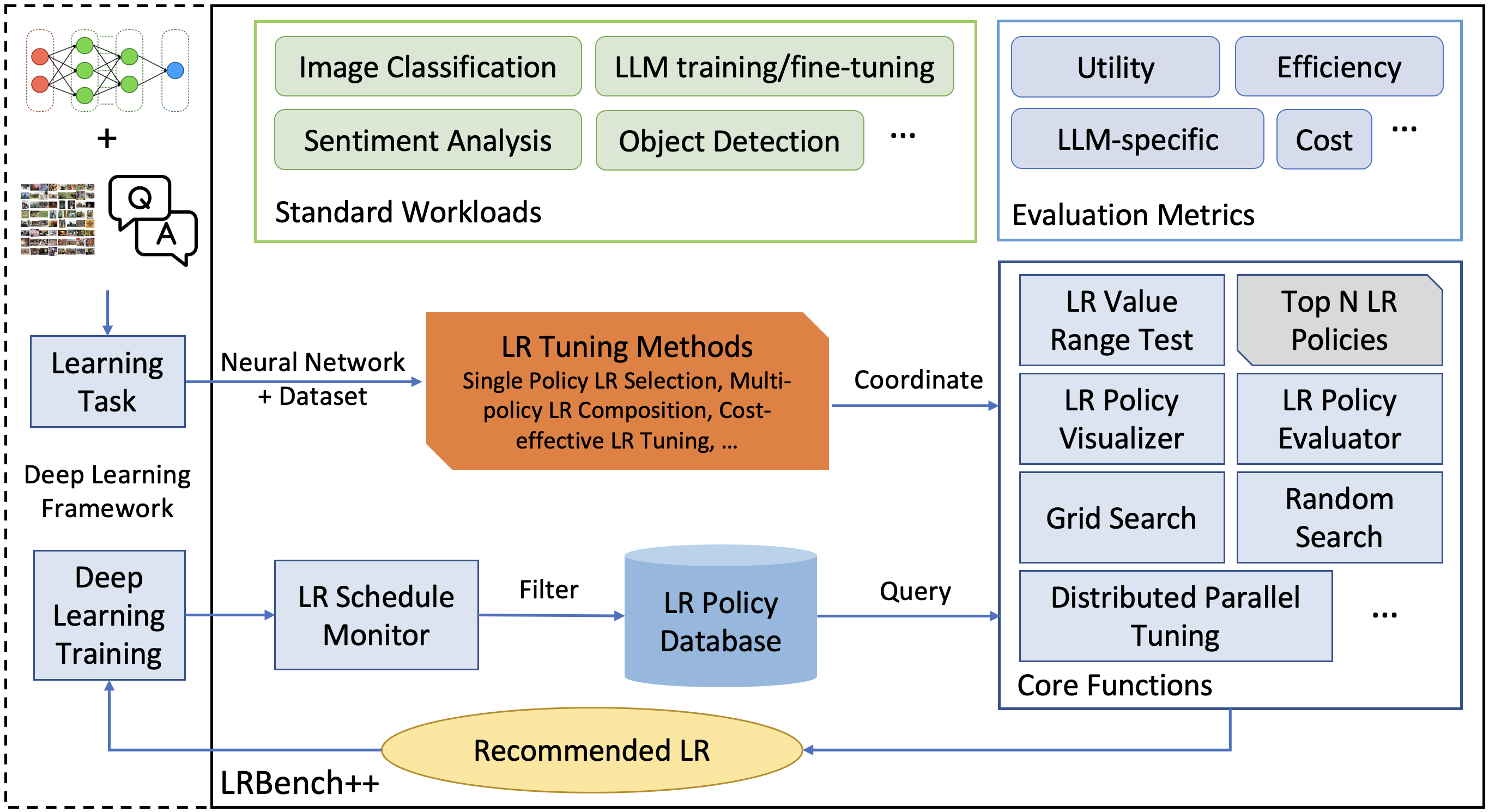}
\caption{Overview of Core Components of LRBench++}
\label{fig:lrbench++-overview}
\end{figure}

The core components of LRBench++ are illustrated in Figure~\ref{fig:lrbench++-overview}, including (1) the LR schedule monitor, which keeps tracking the DNN/LLM training/fine-tuning status and updates the LR value according to a specific LR policy, (2) the LR policy database, which stores the LR tuning results that are organized by the specific learning task, (3) the LR value range test, which leverages the grid search to determine an appropriate LR value range to narrow down the LR search space, (4) the LR policy visualizer, which assists end-users to visualize the DNN/LLM training/fine-tuning status and LR values, (5) the LR policy evaluator, which estimates good LR parameters through the evaluation and comparison of alternative LR policies and dynamic LR tuning, and (6) the LR tuning optimizations, which includes grid search, random search, and distributed parallel tuning that are built on top of Ray Tune~\cite{ray-tune}.
We provide several core features in LRBench++ to effectively support LLM training/fine-tuning, such as (1)~representative LLM implementations, such as LLaMA~\cite{llama} and Alpaca~\cite{alpaca}, (2) iteration-based LR tuning, which is critical for LLM use cases often with only 2$\sim$3 epochs and provides more fine-grained tuning than the epoch-based tuning, and (3) systematic NLP learning tasks for evaluating LLMs, such as ARC~\cite{arc}, MMLU~\cite{mmlu}, HellaSwag~\cite{hellaswag}, and TruthfulQA~\cite{truthfulqa}.

LRBench++ adopts a flexible modular design to enable users to utilize different modules to perform learning rate benchmarking and/or tuning.
For benchmarking LRs, LRBench++ can be used for different deep learning tasks, deep learning frameworks, and hardware devices, which allow LRBench++ to provide benchmark results under various real-world scenarios.
For tuning LRs, LRBench++ supports three LR tuning methods with different tuning objectives: (1) single policy LR selection, (2) multi-policy LR composition, and (3) cost-effective LR tuning. The first and second methods aim to maximize the model accuracy under the given training/fine-tuning constraints, such as a pre-defined number of training/fine-tuning iterations. The third method aims at minimizing the training costs while achieving a target accuracy threshold.
We defer the detailed descriptions of how to use LRBench++ to Section~\ref{section:experimental-analysis}, where we will introduce LRBench++ functionalities under specific scenarios.

We envision that LRBench++ can serve as a public platform for end-users to collect and share learning rate tuning results and facilitate effective learning rate tuning for both traditional DNNs and LLMs. 
LRBench++ is available as an open-source project on GitHub (\url{https://github.com/mlsysx/LRBenchPlusPlus}).

\section{Experimental Analysis} \label{section:experimental-analysis}
We conduct the experiments using the Nvidia RTX 3090 GPU for evaluating traditional DNNs, and 8 Nvidia A100 GPUs for fine-tuning LLMs.
We evaluate LR policies by training two popular DNNs, LeNet~\cite{mnistlenet} and ResNet~\cite{resnet}, on three benchmark datasets, MNIST~\cite{mnistlenet}, CIFAR-10~\cite{cifar10-100}, and ImageNet~\cite{ILSVRC}, and fine-tune an LLM, LLaMA-7B, on 52K instructions following~\cite{alpaca}.
For the most critical experiment results, we report the mean$\pm$std values after 5 repeats.
We evaluate fine-tuned LLMs with four representative tasks, including (1) ARC (AI2 Reasoning Challenge, 25-shot)~\cite{arc}, which is a set of multi-choice questions selected from 3rd to 9th-grade science exams, (2) HellaSwag (10-shot)~\cite{hellaswag}, which is a test of commonsense inference that is challenging for language models but easy for humans, (3) MMLU (2-shot)~\cite{mmlu}, which is a massive multitask test that requires models to possess extensive world knowledge and problem-solving abilities to achieve high accuracy, and (4) TruthfulQA (0-shot)~\cite{truthfulqa}, which is a benchmark that evaluates truthfulness of answers generated by the model. We refer readers for the definitions of LR policies and configurations to~\cite{LRBenchBigData,LRBenchTIST}.

\begin{table}[h!]
\caption{Accuracy Comparison of 12 Learning Rate Policies for training ResNet32 on CIFAR-10}
\label{table:lr-comparison-cifar10-resnet}
\centering
\scalebox{0.7}{
\small
\begin{tabular}{c|cccccc|c}
\hline
\multicolumn{7}{c|}{Learning Rate Policy}  & \multirow{2}{*}{Accuracy (\%)}\\ \cline{1-7}
Category & Function & $k (k_0)$  & $k_1$    & $\gamma$   & $p$    & $l$  &  \\ \hline
\multirow{4}{*}{Fixed LR}    & FIX         & 0.1    &     &         &      &              & 86.08         \\
                             & FIX         & 0.01   &     &         &      &              & 85.41         \\
                             & FIX         & 0.001  &     &         &      &              & 82.66         \\
                             & FIX         & 0.0001 &     &         &      &              & 66.96         \\ \hline
\multirow{4}{*}{Decaying LR} & \textbf{NSTEP}       & 0.1    &     & 0.1     &      & \footnotesize{32000, 48000} & \textbf{92.38$\pm$0.04}   \\
                             & STEP        & 0.1    &     & 0.99994 &      &              & 91.10         \\
                             & EXP         & 0.1    &     & 0.85    &      & 5000         & 91.03         \\
                             & \textbf{POLY}        & 0.1    &     &         & 1.2  &              & \textbf{92.39}         \\ \hline
\multirow{6}{*}{Cyclic LR}   & TRI         & 0.0001 & 0.9 &         &      & 2000         & 76.91         \\
                             & TRI2        & 0.0001 & 0.9 &         &      & 2000         & 91.85         \\
                             & \textbf{TRIEXP}      & 0.0001 & 0.9 & 0.99994 &      & 2000         & \textbf{92.76$\pm$0.14}   \\
                             & SIN         & 0.0001 & 0.9 &         &      & 2000         & 72.78         \\
                             & SIN2        & 0.0001 & 0.9 &         &      & 2000         & 91.98         \\
                             & \textbf{SINEXP}      & 0.0001 & 0.9 & 0.99994 &      & 2000         & \textbf{92.81$\pm$0.08}   \\ \hline
\multirow{3}{*}{\begin{tabular}[c]{@{}c@{}}Advanced\\ Composite LR\end{tabular}} & \multicolumn{6}{c|}{\textbf{MULTI}: 0-30000 iter:TRI ($k_0$=0.1, $k_1$=0.5, $l$=1500)} & \multirow{3}{*}{\textbf{92.91\%}} \\ 
                          & \multicolumn{6}{c|}{30000-60000 iter: TRI ($k_0$=0.01, $k_1$=0.05, $l$=1000)}  &  \\ 
                          & \multicolumn{6}{c|}{60000-64000 iter: TRI ($k_0$=0.001, $k_1$=0.005, $l$=500)} &  \\ \hline
\end{tabular}
} 
\end{table}

\begin{figure}[h!]
\centering
\subfloat[Cyclic LR (SINEXP): 92.81\% Acc]{
  \centering
  \includegraphics[width=0.245\textwidth]{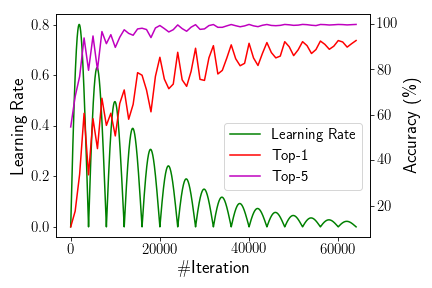}
  \label{fig:sinexp-cifar10-resnet32}
} 
\subfloat[Multi-policy LR: 92.91\% Acc]{
  \centering
  \includegraphics[width=0.245\textwidth]{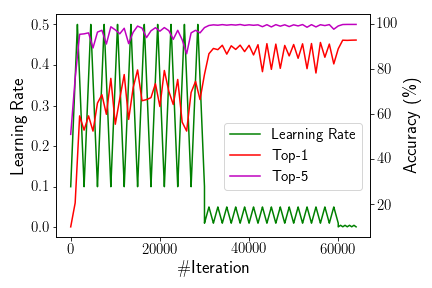}
  \label{fig:hlr-cifar10-resnet32}
} 
\caption{Comparison of SINEXP and Multi-policy LRs (CIFAR-10, ResNet32), showing the benefit of advanced composite LR in improving DNN accuracy}
\label{fig:lr-comparison-cifar10-resnet32}
\end{figure}

\noindent \textbf{Accuracy Comparison of LR Policies.}
We leverage LRBench++ to compare various LR policies on CIFAR-10 and ImageNet datasets and highlight the impacts of effective LR policies in improving the DNN model accuracy.
Table~\ref{table:lr-comparison-cifar10-resnet} presents the results of training ResNet32 on CIFAR-10 using LRBench++ with the default batch size of 128 and 64,000 training iterations. We highlight five interesting observations.
{\it First}, for the baseline fixed LR policies, the model accuracy highly depends on the specific $k$ values. For instance, with $k$=0.0001, the ResNet32 can only achieve 66.96\% accuracy, which is much lower than most other LR policies with over 91\% accuracy. Therefore, it is challenging to select a proper fixed LR value for training DNNs.
{\it Second}, all decaying LR policies can train ResNet32 with high accuracy of over 91.03\%. LRBench++ recommends these decaying LR policies based on the optimal $k$=0.1 obtained during the fixed LR tuning, which shows that such a selection of initial LR values is effective.
{\it Third}, two cyclic LR policies, TRIEXP (92.76\%) and SINEXP (92.81\%), achieve much higher accuracy than the best decaying LR policy, POLY (92.39\%), showing that cyclic LR policies can further improve DNN accuracy.
{\it Fourth}, the advanced composite LR applies three TRI functions during DNN training and achieves the highest accuracy of 92.91\%, which outperforms the best cyclic LR policy, SINEXP (92.81\%), and demonstrates the high potential of multi-policy LRs in improving DNN accuracy. 
{\it Fifth}, LRBench++ can effectively evaluate the performance of various LR policies and identify the top-performing LR policies. For example, the five LR policies in bold in Table~\ref{table:lr-comparison-cifar10-resnet} (NSTEP (92.38\%), POLY (92.39\%), TRIEXP (92.76\%), SINEXP (92.81\%), and MULTI (92.91\%)) will be recommended by LRBench++ as the Top-5 choices, which can achieve over 92.38\% accuracy and improve the best baseline fixed LR ($k$=0.1 with 86.08\% accuracy) by over 6.3\%.

We further visualize the training process in terms of the LR values (green) and Top-1/Top-5 accuracy (red/purple) in Figure~\ref{fig:lr-comparison-cifar10-resnet32} for the two best performing LRs, SINEXP and the advanced composite LR, in Table~\ref{table:lr-comparison-cifar10-resnet}. We highlight two interesting observations.
{\it First,} the Top-1 and Top-5 accuracy increases rapidly at the beginning, which indicates that starting the DNN training by increasing LRs to relatively large values can be beneficial.
{\it Second,} at the end of the training, SINEXP still has higher LR values than this composite LR, which slows down the model convergence and results in lower model accuracy of 92.81\%. In contrast, the composite LR leverages three different cyclic LRs, which can effectively adapt to different training stages and achieve the highest accuracy of 92.91\%.

Table~\ref{table:hclr-comparison-imagenet} presents the experimental results for training ResNet18 on another dataset, ImageNet. We found similar observations. The advanced composite LR policy consisting of three FIX LR functions achieves the highest 69.05\% Top-1 accuracy and 88.76\% Top-5 accuracy, followed by the decaying LR, STEP, and fixed LR.

\begin{table}[h!]
\centering
\caption{Accuracy Comparison of 3 Learning Rate Policies for training ResNet18 on ImageNet}
\label{table:hclr-comparison-imagenet}
\small
\scalebox{0.7}{
\begin{tabular}{c|c|cc}
\hline
\multicolumn{2}{c|}{LR Policy} &
  \multirow{2}{*}{Top-1 (\%)} &
  \multirow{2}{*}{Top-5 (\%)} \\ \cline{1-2}
Category        & LR Function                                                 &       &       \\ \hline
Fixed LR    & FIX ($k$=0.1)                                                 & 50.50 & 76.51 \\ \hline
Decaying LR & STEP ($k$=0.1, $\gamma$=0.1, $l$=30 epoch)                       & 68.40 & 88.37 \\ \hline
\multirow{4}{*}{\begin{tabular}[c]{@{}c@{}}Advanced\\Composite LR\end{tabular}} &
  0-5 epoch: FIX ($k$=0.1) &
  \multirow{4}{*}{\textbf{69.05}} &
  \multirow{4}{*}{\textbf{88.76}} \\ 
            & \makecell{5-35 epoch: SINEXP\\($k_0$=0.1, $k_1$=0.6, $l$=15 epoch, $\gamma$=0.999987)}    &       &       \\ 
            & 35-50 epoch: FIX ($k$=0.01)                                       &       &       \\ 
            & 50-60 epoch: FIX ($k$=0.001)                                      &       &      \\ \hline
\end{tabular}
} 
\end{table}

\noindent \textbf{LR Tuning for Cost Reduction.}
For some scenarios or deep learning applications, such as the pruning in neural network search~\cite{neuralnetworksearch}, the objective of training/fine-tuning is to achieve a target accuracy threshold and minimize the training/fine-tuning time cost, i.e., the number of training/fine-tuning iterations. With a target accuracy threshold, LRBench++ can recommend effective LR policies, monitor the training/fine-tuning process, and stop the training/fine-tuning to reduce time costs when the target accuracy is met. Table~\ref{table:lr-cost-comparison} shows the minimum \#Iterations for achieving the target accuracy during training DNNs on MNIST and CIFAR-10. We highlight two observations.
{\it First}, it is challenging to achieve the target accuracy using the fixed LR policies. For MNIST, only one single fixed LR with $k$=0.01 is able to reach the target accuracy of 99.1\% at the end of the entire 10,000 training iterations. All baseline fixed LR policies (see Table~\ref{table:lr-comparison-cifar10-resnet}) fail to reach the target accuracy of 90\% for CIFAR-10.
{\it Second,} cyclic LR policies can reach the target accuracy threshold with a reduced cost. For MNIST, the SIN2 can reduce the entire cost of 10,000 training iterations to 3,500 by 2.86$\times$. For CIFAR-10, the two cyclic LR policies, TRI2 and SIN2, achieve the target accuracy of 90\% on CIFAR-10 with only 20,000 iterations and significantly reduce the default 64,000 iterations by 3.2$\times$.

\begin{table}[h!]
\caption{Cost Comparison of Learning Rate Policies for Achieving Target Accuracy}
\label{table:lr-cost-comparison}
\centering
\scalebox{0.7}{
\small
\begin{tabular}{cccc}
\hline
Dataset \& Model                                                                                  & Learning Rate Policy                                           & \makecell{\#Iter @\\ Target Acc}  & Speedup \\ \hline
\multirow{4}{*}{\begin{tabular}[c]{@{}c@{}}MNIST (LeNet,\\Default:10,000 iters\\Target Acc=99.1\%)\end{tabular}}   & FIX (k=0.01)                                                       & 10000 & 1.00$\times$ \\ \cline{2-4}
& \makecell{NSTEP ($k$=0.01, $\gamma$=0.9,\\$l$=[5000, 7000, 8000, 9000])} & 10000 & 1.00$\times$ \\ \cline{2-4}
& POLY($k$=0.01, $p$=1.2)                                     & 7000  & \textbf{1.43$\times$}  \\ \cline{2-4}
& SIN2 ($k_0$=0.01, $k_1$=0.06, $l$=2000)               & \textbf{3500} & \textbf{2.86$\times$} \\ \hline
\multirow{4}{*}{\begin{tabular}[c]{@{}c@{}}CIFAR-10 (ResNet32,\\Default:64,000 iters\\Target Acc=90\%)\end{tabular}} & NSTEP ($k$=0.1, $\gamma$=0.1, $l$=[32000, 48000])           & 33000 &1.94$\times$               \\ \cline{2-4}
& POLY ($k$=0.1, $p$=1.2)                             & 51000  &1.25$\times$             \\ \cline{2-4}
& TRI2 ($k_0$=0.0001, $k_1$=0.9, $l$=2000)            & \textbf{20000} &\textbf{3.2$\times$}  \\ \cline{2-4}
& SIN2 ($k_0$=0.0001, $k_1$=0.9, $l$=2000)            & \textbf{20000} & \textbf{3.2$\times$}          \\ \hline     
\end{tabular}
} 
\end{table}

\noindent \textbf{LR Tuning by Grid Search.}
It is challenging to perform LR tuning by using grid search given numerous LR functions and multiple LR parameters (e.g., $\ge$3 for cyclic LRs), which may require a large high-dimensional search space.
Some empirical constraints can be introduced to effectively reduce the dimensionality of the search space. For example, the LR coefficient $\lambda$ can control the LR value scale. By setting the actual LR value $\eta'(t) = \lambda \eta(t)$, we can perform the grid search on $\eta(t)$ and $\lambda$ using only two dimensions to identify the Top-$K$ LR policies.

\begin{figure}[t!]
    \centering
    \includegraphics[width=0.4\textwidth]{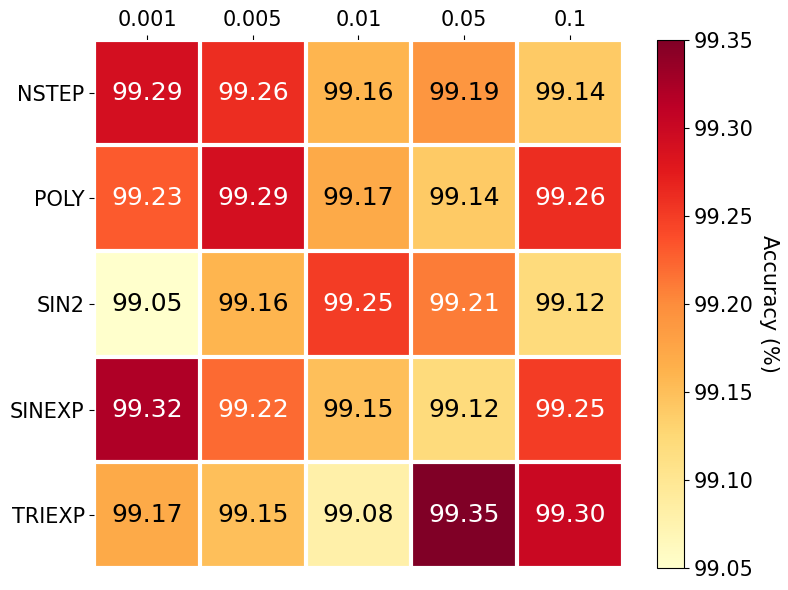}
\caption{Learning Rate Tuning by Grid Search (MNIST)}
\label{fig:lr-tuning-grid-search-mnist}
\end{figure}

We evaluate the effectiveness of the learning rate tuning via grid search in LRBench++ for training LeNet on MNIST with 20 epochs. We compare five LR policies, NSTEP($k=1.0, \gamma=0.9, l=[10, 14, 16, 18, 19]$), POLY($k=1.0,p=1.2$) and three cyclic LRs, SIN2, SINEXP and TRIEXP sharing $k_0=1, k_1=3,l=5$ with $\gamma=0.94$ for both SINEXP and TRIEXP. The LR coefficient $\lambda$ is used to reduce the search space dimension.
Figure~\ref{fig:lr-tuning-grid-search-mnist} shows the LR tuning results, where the x-axis represents the LR coefficients $\lambda$, the y-axis marks the five LR policies, and the darker the color indicates the higher the accuracy. We highlight three interesting observations.
{\it First}, the different combinations of the LR coefficients and LR policies will result in dramatic differences in the trained model accuracy. For some combos, such as SIN2 with $\lambda$=0.001 and TRIEXP with $\lambda$=0.01, they fail to deliver the target accuracy of 99.10\% while a different combo with the same LR policy or the same LR coefficient can produce high accuracy of over 99.25\%, such as SIN2 with $\lambda$=0.01 and TRIEXP with $\lambda$=0.05, which shows that it is a non-trivial task to identify the good LR policies to achieve high accuracy DNN training.
{\it Second}, LRBench++ can help effectively identify good LR policies through grid search. For example, TRIEXP with $\lambda$=0.05 achieves the highest accuracy of 99.35\%. The other Top-$K$ (e.g., $K$=5) LR policies will also be recommended by LRBench++ for training LeNet on MNIST.

LRBench++ can further improve or verify the LR performance by using grid search. After determining the optimal combo of LR policy and coefficient, such as TRIEXP with $\lambda$=0.05, LRBench++ can continue the grid search for other LR parameters based on user configurations. For example, the ratio of $k_1$ and $k_0$, i.e., $k_1/k_0$ for TRIEXP, may also impact the performance of TRIEXP. We leveraged grid search in LRBench++ to vary $k_1/k_0$ from 2 to 6 and obtained the trained model accuracy of 99.26\% ($k_1/k_0$=2), \textbf{99.35\%} ($k_1/k_0$=3), 99.10\% ($k_1/k_0$=4), 99.14\% ($k_1/k_0$=5), and 99.25\% ($k_1/k_0$=6), where $k_1/k_0$=3 is still the optimal ratio for achieving the highest accuracy for TRIEXP. This process can be continued to tune other LR parameters for TRIEXP, such as $l$ and $\gamma$.

\noindent \textbf{Impact of LR Tuning Algorithms.}
We provide the support of different LR tuning algorithms in LRBench++, including (1) random search, which randomly selects the hyperparameters within a pre-defined range of parameter values, such as $[0, 1]$, and then runs the DNN training over randomly picked hyperparameter values and (2) grid search, where users will feed LRBench++ with their choice of hyperparameter values and use these values to perform DNN/LLM training/fine-tuning and collect results.
We compare grid search and random search by training LeNet on MNIST with LRBench++. We repeated the experiments 5 times and report the mean±std for model accuracy in Table~\ref{table:comparison-grid-random-search-mnist}. We utilize the LR coefficient $\lambda$ and TRIEXP($k_0=1, k_1=3,\gamma=0.94,l=5$) to reduce the search space dimension.
Random search is set to select $N$=5 number of $\lambda$ in the continuous range of $[0.001, 0.1]$, and grid search is set to $N$=5 number of $\lambda$ in the same range but manually selected by us based on Top-$K$ LRBench++ recommendations. As a result, the LR selection by grid search delivered better accuracy results than the choice by random search with a lower standard deviation.

\begin{table}[h!]
\centering
\caption{Comparison of Grid Search and Random Search}
\label{table:comparison-grid-random-search-mnist}
\small
\scalebox{0.72}{
\begin{tabular}{ccc}
\hline
Method         & $\lambda$ values & Accuracy   (\%) \\ \hline
Grid   Search  & 0.001, 0.005, 0.01, 0.05, 0.1 & 99.30$\pm$0.04      \\ \hline
Random   Search & random from [0.001, 0.1] & 99.17$\pm$0.06     \\ \hline
\end{tabular}
} 
\end{table}

\begin{table}[h!]
\centering
\caption{The evaluation results of each model in each epoch}
\label{table:llm-comparison-lr-policy}
\renewcommand{\arraystretch}{1.5}
\scalebox{0.72}{
\small
\begin{tabular}{cc|cccc}
\hline
\multicolumn{2}{c|}{Model} & \multicolumn{4}{c}{Evaluation} \\
\hline
Model Type & Stage & ARC(25) & HellaSwag(10) & MMLU(2) & TruthfulQA(0) \\
\hline
Pre-trained LLM & Baseline & 0.5119 & 0.7776 & 0.3414 & 0.3407 \\
\hline

\multirow{3}{*}{2e-05 (linear)} & 1 Epoch & 0.5162 & 0.7599 & 0.3691 & 0.3874 \\
                       & 2 Epoch & \textbf{0.5392} & \textbf{0.7843} & 0.4021 & 0.3981 \\
                       & 3 Epoch & 0.5282 & 0.7821 & \textbf{0.4110} & \textbf{0.4018} \\
\hline
\multirow{3}{*}{6e-05 (linear)} & 1 Epoch & 0.4599 & 0.7077 & 0.2944 & 0.4072 \\
                       & 2 Epoch & \textbf{0.4667} & \textbf{0.7305} & \textbf{0.3556} & \textbf{0.4113} \\
                       & 3 Epoch & 0.4471 & 0.7079 & 0.3312 & 0.3948 \\
\hline
\multirow{3}{*}{2e-05 (cosine)} & 1 Epoch & 0.5077 & 0.7573 & 0.3704 & 0.3889 \\
                       & 2 Epoch & \textbf{0.5350} & \textbf{0.7824} & 0.4017 & 0.3944 \\
                       & 3 Epoch & 0.5247 & 0.7797 & \textbf{0.4194} & \textbf{0.4036} \\
\hline
\multirow{3}{*}{1e-05 (fixed)} & 1 Epoch & 0.5154 & 0.7684 & 0.3807 & 0.3815 \\
                             & 2 Epoch & \textbf{0.5503} & 0.7875 & 0.3915 & 0.3916 \\
                             & 3 Epoch & 0.5282 & \textbf{0.7919} & \textbf{0.4058} & \textbf{0.3940} \\
\hline
\end{tabular}
}
\end{table}

\noindent \textbf{Impact of LRs on LLM fine-tuning.}
We compare various LR policies on fine-tuning LLaMA-7B on the 52K instruction dataset~\cite{alpaca} and study their impacts on LLM fine-tuning using LRBench++.
Table~\ref{table:llm-comparison-lr-policy} presents the experimental results with the default batch size of 128 and 1218 training iterations. We highlight four interesting observations. 
{\it First}, the learning rate has a high impact on LLM fine-tuning outcomes.
In Table~\ref{table:llm-comparison-lr-policy}, we observe that different LR values can lead to highly different NLP task performance for fine-tuning LLaMA-7B. For example, when evaluated with the ARC (AI2 Reasoning Challenge)~\cite{arc}, the linear decaying LR with the initial LR value of $2 \times 10^{-5}$ can produce 0.5392 ARC score compared to 0.4667 by the linear decaying LR of $6 \times 10^{-5}$ for achieving their optimal ARC scores with 2 epochs. In particular, the optimal 0.4667 ARC score produced by the linear decaying LR of $6 \times 10^{-5}$ is even lower than the baseline pre-trained LLM before fine-tuning.
{\it Second,} the learning rate tuning can potentially enhance LLM fine-tuning efficiency. In Table~\ref{table:llm-comparison-lr-policy}, the best performing model may not require the full 3 epoch fine-tuning, such as 2 epochs for the linear decaying LR with the initial LR value of $2 \times 10^{-5}$ under the ARC test. Given the high cost of LLM training/fine-tuning, effective learning rate tuning can potentially lead to huge training/fine-tuning cost savings.
{\it Third,} no single model can outperform the others in more than one test. For the ARC test, the best performing model is the LLM fine-tuned with a fixed learning rate of $1 \times 10^{-5}$ for 2 epochs. For HellaSwag, the best LLM is produced by the same LR with 3 epoch fine-tuning. The LLM fine-tuned with the cosine decaying LR with the initial LR value of $2 \times 10^{-5}$ wins the MMLU test after 3 epoch fine-tuning, while the winner for TruthfulQA is the LLM fine-tuned with the linear decaying LR of $6 \times 10^{-5}$ for 2 epochs.
{\it Fourth}, the fixed learning rate with a constant LR value of $1 \times 10^{-5}$ can produce a high quality LLM with the best performance on both ARC and HellaSwag tests, outperforming the other linear and cosine decaying LRs. This shows that the fixed LRs work well for fine-tuning LLMs, challenging the prevailing perception that fixed LRs are often less effective in deep learning training/fine-tuning~\cite{LRBenchBigData,clr,sgdr}.
These observations demonstrate the varying requirements for LLM training/fine-tuning with different hyperparameter configurations and evaluation tasks.

\begin{figure}[h!]
\centering
\subfloat[Learning rate]{
  \centering
  \includegraphics[width=0.2\textwidth]{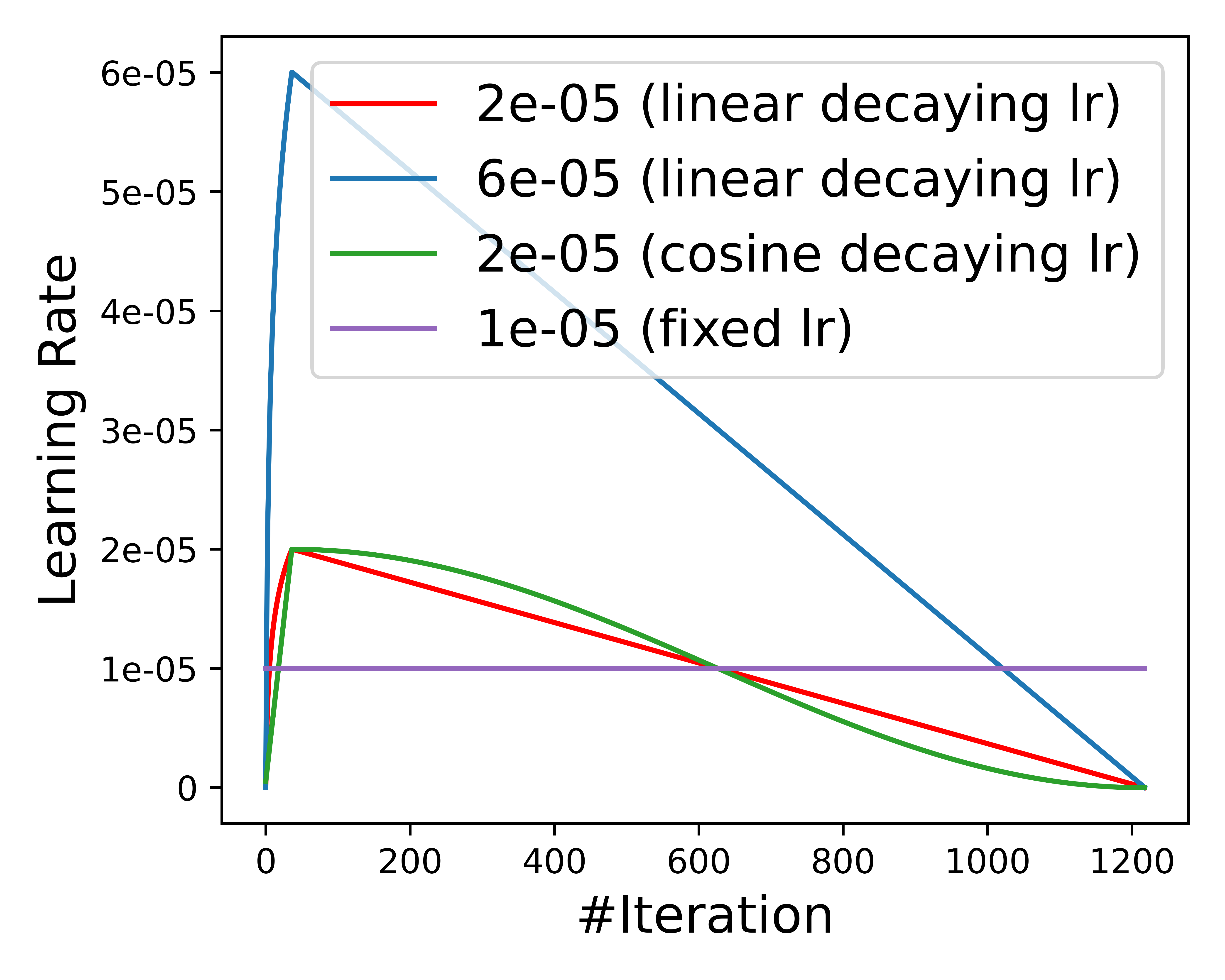}
  \label{fig:llm-learning-rate}
} 
\subfloat[Training loss]{
  \centering
  \includegraphics[width=0.24\textwidth]{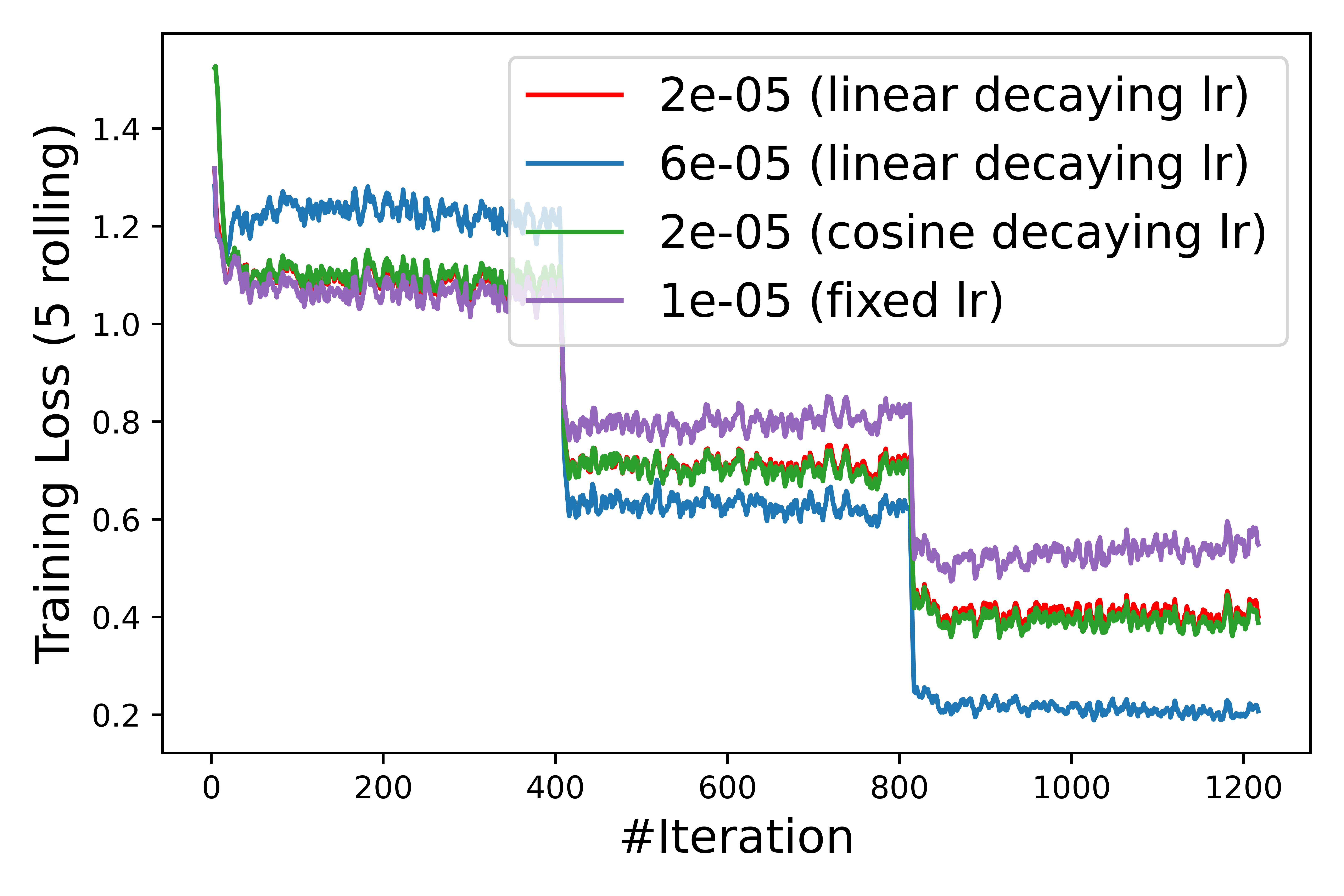}
  \label{fig:llm-training-loss}
} 
\caption{The left figure illustrates the learning rate values under three different LR policies throughout the LLM fine-tuning process. The right figure displays the corresponding training loss, smoothed using a 5-point rolling average.}
\label{fig:llm-finetuning-comparasion}
\end{figure}
We further visualize the LLM fine-tuning process in terms of the LR values in Figure~\ref{fig:llm-learning-rate} for the LLMs fine-tuned with the fixed LR of $1 \times 10^{-5}$ (purple), the cosine decaying LR of $2 \times 10^{-5}$ (green), and two different linear decaying LRs of $2 \times 10^{-5}$ (red) and $6 \times 10^{-5}$ (blue), and their training loss with the corresponding colors in Figure~\ref{fig:llm-training-loss}. 
We highlight three interesting observations.
{\it First,} similar LR values can produce similar LLM fine-tuning outcomes. Figure~\ref{fig:llm-learning-rate} shows that the LR values of the linear decaying LR of $2 \times 10^{-5}$ (red) and the cosine decaying LR of $2 \times 10^{-5}$ (green) are very close, leading to similar training loss in Figure~\ref{fig:llm-training-loss} as well as similar task-specific performance in Table~\ref{table:llm-comparison-lr-policy}.
{\it Second,} the training loss may not accurately reflect the LLM task-specific performance. For example, during the 2nd epoch, the fixed LR of $2 \times 10^{-5}$ (purple) produces the highest training loss while achieving the best score on the ARC test in Table~\ref{table:llm-comparison-lr-policy}.
{\it Third,} the trend of the training loss reveals a unique stair-step decline pattern after initially decreasing during the early iterations, which is consistent with~\cite{finetuning-llama2}. The significant loss drops at epochs 2 and 3 explain the enhanced model performance observed in Table~\ref{table:llm-comparison-lr-policy}. Moreover, this pattern also suggests that fine-tuning LLMs may exhibit different characteristics from LLM training~\cite{llama,llama2}, making it a unique research challenge to investigate. 

To summarize our experimental analysis, both the learning rate and the choice of learning rate policies have significant impacts on the DNN model training and LLM fine-tuning performance, which provide potential opportunities and require further studies for optimizing DNN/LLM training/fine-tuning performance.

\section{Discussion}

In this section, we first summarize the common practices of LLM fine-tuning settings and then discuss the research challenges and opportunities of learning rate tuning in the era of Large Language Models.

\begin{table}[h]
\caption{Common LLM Fine-tuning Settings in Practice}
\label{table:fine-tuning-method-comparasion}
\centering
\scalebox{1.0}{
    \begin{tabular}{p{2.2cm}|p{0.8cm}p{1.6cm}rp{0.8cm}}
        \hline
        \multirow{2}{*}{\parbox{2.2cm}{\centering Model}}
 & Learning Rate & LR Scheduler \newline (DeepSpeed) & Epoch & Warmup \newline \% or \# \\
        \hline
        Alpaca-7B~\cite{alpaca} & 2e-5 & WarmupDecayLR & 3 & 0.03 \\
        Vicuna-7B~\cite{vicuna} & 2e-5 & cosine & 3 & 0.03 \\
        WizardLM-7B~\cite{wizardlm} & 2e-5 & cosine & 3 & 2 \\
        GPT4All-J-6B~\cite{gpt4all} & 2e-5 & WarmupLR & 2 & 500 \\
        Dolly-7B~\cite{dolly} & 5e-6 & WarmupLR & 2 & 50 \\
        \hline
    \end{tabular}
} 
\end{table} 

\noindent \textbf{Common LR Settings for LLM Fine-tuning.} We review and summarize the popular LLM fine-tuning practices in Table~\ref{table:fine-tuning-method-comparasion}. Most existing LLM fine-tuning studies simply leverage the default initial LR value of $2 \times 10^{-5}$ and the default LR policy, such as the cosine decaying LR or WarmupLR~\cite{sgdr,deepspeed,alpaca}, which may not achieve the optimal LLM fine-tuning results. For example, Table~\ref{table:llm-comparison-lr-policy} shows that simply using a fixed LR of $2 \times 10^{-5}$ can achieve the best performance on both ARC and HellaSwag, which also outperforms the popular cosine decaying LR of $2 \times 10^{-5}$ adopted by Vicuna-7B~\cite{vicuna} and  WizardLM-7B~\cite{wizardlm} in Table~\ref{table:fine-tuning-method-comparasion}. This further confirms our previous analysis and demonstrates the pressing need to rethink learning rate tuning for both DNN and LLM training/fine-tuning.

\noindent \textbf{Challenges and Opportunities in LR Tuning.} We below discuss the challenges and opportunities in LR tuning from three perspectives.
(1) \textbf{Cost-effective LR tuning:} given the substantial computational costs of LLM training/fine-tuning, it is both challenging and costly to perform an exhaustive search of potential LR policies for a new LLM or a new learning task. From a practical perspective, it presents an attractive research opportunity to identify effective learning rate policies or tuning strategies that can efficiently achieve desired model accuracy within a given budget.
(2) \textbf{Benchmarking LR}: this challenge lies in how to accurately capture the LR impacts on DNN/LLM training/fine-tuning in terms of different hyperparameter configurations, costs, and trained/fine-tuned model quality.
(3) \textbf{LLM Performance Evaluation during Training/Fine-tuning}: in Section~\ref{section:experimental-analysis}, we found that the training/validation loss during the LLM fine-tuning may not effectively reflect the LLM task-specific performance, such as ARC or HellaSwag tests. This suggests that the state-based LRs that rely on monitoring the training/validation loss, such as Reduce-LR-On-Plateau~\cite{reducelronplateau} and Change-LR-On-Plateau~\cite{LRBenchTIST}, may not work well for LLM training/fine-tuning. It will be challenging and valuable to study how to evaluate LLM task-specific performance during LLM training/fine-tuning to obtain timely feedback for LR tuning.

\section{Conclusion}
We revisit the changes from traditional DNNs to Large Language Models in fundamental assumptions and identify the core research challenges and opportunities in learning rate tuning through an empirical study.
This paper makes three original contributions.
{\it First,} we examine existing learning rate policies with the goal of improving traditional deep learning training and LLM fine-tuning efficiency.
{\it Second,} we present LRBench++, a learning rate benchmarking and tuning tool, which can effectively identify good learning rate policies and support LLM fine-tuning.
{\it Third,} our experimental evaluation using LRBench++ on both traditional DNN training and LLM fine-tuning demonstrates the pressing need to investigate learning rate tuning in the era of LLMs and validates our analysis.
We conjecture that this study will inspire in-depth understanding and provide practical guidelines toward efficient DNN/LLM training and fine-tuning.

\bibliographystyle{IEEEtran}
\bibliography{reference}
\end{document}